\pdfoutput=1
\documentclass[11pt]{article}

\usepackage[final]{acl}

\usepackage{times}
\usepackage{latexsym}

\usepackage[T1]{fontenc}
\usepackage[utf8]{inputenc}

\usepackage{microtype}

\usepackage{inconsolata}

\usepackage{graphicx}
\usepackage{paralist}
\usepackage{booktabs}
\usepackage{amsmath,amsfonts}
\usepackage{color} % [usenames,dvipsnames]

\title{
Dissecting Fine-Tuning Unlearning in Large Language Models
}

\author{Yihuai Hong\textsuperscript{1,4}\footnotemark[1],~Yuelin Zou\textsuperscript{2},~Lijie Hu\textsuperscript{3},~Ziqian Zeng\textsuperscript{1},~Di Wang\textsuperscript{3}\footnotemark[2],~Haiqin Yang\textsuperscript{4}\footnotemark[2]   \\[6pt]%\\
        \textsuperscript{1}South China University of Technology, \ \textsuperscript{2}Columbia University\\ \  \textsuperscript{3}King Abdullah University of Science and Technology  \\ \textsuperscript{4}International Digital Economy Academy (IDEA), China  \\  {\texttt{yihuaihong@gmail.com}, \texttt{hqyang@ieee.org}        %\textbf{Correspondence:} 
        }}

\begin{document}
\maketitle
% \show\thefootnote
\renewcommand{\thefootnote}{\fnsymbol{footnote}}

\footnotetext[1]{Work done during an internship at IDEA.}
\footnotetext[2]{Corresponding authors.}
\begin{abstract}

Fine-tuning-based unlearning methods prevail for preventing targeted harmful, sensitive, or copyrighted information within large language models while preserving overall capabilities. However, the true effectiveness of these methods is unclear. In this work, we delve into the limitations of fine-tuning-based unlearning through activation patching and parameter restoration experiments. Our findings reveal that these methods alter the model's knowledge retrieval process, providing further evidence that they do not genuinely erase the problematic knowledge embedded in the model parameters. Instead, the coefficients generated by the MLP components in the model's final layer are the primary contributors to these seemingly positive unlearning effects, playing a crucial role in controlling the model's behaviors. Furthermore, behavioral tests demonstrate that this unlearning mechanism inevitably impacts the global behavior of the models, affecting unrelated knowledge or capabilities. The code is released at \url{https://github.com/yihuaihong/Dissecting-FT-Unlearning}.

%This study provides critical insights into the constraints of fine-tuning-based unlearning, highlighting the need for more comprehensive techniques to truly and selectively remove unwanted knowledge from large language models without compromising their overall performance.

\end{abstract}

\section{Introduction}
Large language models (LLMs), due to their extensive pre-training corpora, often inadvertently learn harmful, sensitive, or copyright-protected knowledge~\citep{chang2023speak, mozes2023use, eldan2023whos, ye2022learning}. Consequently, recent research has focused on developing efficient unlearning methods as a post-training technique to selectively unlearn the specific knowledge~\citep{blancojusticia2024digital, liu2024rethinking}. Currently, the core mechanism of these unlearning methods involves finetuning~\citep{eldan2023whos, jang-etal-2023-knowledge, yao2024large, rafailov2023direct}, with corresponding adjustments and designs in the loss function to facilitate the unlearning process.  %\blue{The fine-tuning-based methods aim to modify the model's parameters and internal representations in a targeted manner, to unlearn the unwanted knowledge while preserving the model's overall capabilities.}
Although earlier investigations~\citep{hong2024intrinsicevaluationunlearningusing, lee2024a} have proven that these methods are ineffective at completely erasing model-embedded knowledge, the factors contributing to the misleading success of these techniques remain unclear.

Therefore, in this paper, we try to unveil why existing finetuning-based unlearning methods perform well in behavioral tests by analyzing the mechanisms of internal knowledge recall and flow within models~\citep{meng2022locating, pochinkov2024dissecting, geva-etal-2021-transformer}. %  in \S\ref{sec: patching}
Specifically, we investigate which components or parameters carry these unlearning effects. We design activations patching and parameters restoration experiments in three settings, aiming to independently study the impact of unlearning methods on the coefficients and value vectors in the MLPs, as well as on the attention components' states. Our findings further confirm that the methods do not truly alter the knowledge embedded in the value vectors of MLPs, and reveal that they will change how they extract and transfer this knowledge through modifications in the coefficients of MLPs and attention components during unlearning. 
Notably, the coefficients produced by the MLP in the final layers are primarily responsible for achieving the unlearning effects of finetuning-based methods. 

We further test the global behavior impact of these fine-tuning-based unlearning methods on LLaMA2-7B-chat~\citep{touvron2023llama} and OLMo-7B~\citep{groeneveld2024olmo} by implementing them on the respective pretraining datasets of both models, aiming to more closely simulate the erasure of knowledge acquired during the pretraining process. We discover that while these methods appear to effectively unlearn target knowledge, they also inevitably affect the output and behavior related to unrelated knowledge. This unintended consequence stems from the fact that these approaches are based on altering the model's internal knowledge retrieval mechanisms, thereby impacting its global behavior and overall performance.

Ultimately, we conclude once again that current fine-tuning-based unlearning methods cannot completely erase sensitive knowledge embedded in models, particularly within the MLPs, instead adjusting the mechanisms by which the model retrieves knowledge.
These methods are vulnerable to recovery attacks in components' activations and unsuitable for true unlearning. We advocate for future unlearning evaluations to concentrate on precise measurement of both the actual storage of targeted knowledge within the model's entire parameter set and the specific dynamics of how this knowledge is retrieved and utilized.

% We look forward to the development of stronger parameter modification techniques for unlearning that can completely erase the targeted knowledge.

\section{Background and Related Work}

%\subsection{Unlearning in Large Language Models}
\paragraph{Unlearning in Large Language Models}
Since large language models learn knowledge from different domains and corpora during the pre-training process, it is often found that they contain harmful, sensitive or private knowledge, leading to the possibility that language models produce output behaviors containing corresponding sensitive or harmful information \citep{liu2024rethinking, chang2023speak, mozes2023use}. Therefore, unlearning emerges as a timely and important post-pretraining processing method for LLM safety. Currently, the vast majority of LLM unlearning methods use fine-tuning as the primary operational approach. In terms of classifying them by different training objectives, they include gradient direction control \citep{jang-etal-2023-knowledge, yao2024large, yao2023large} and preference optimization methods \citep{rafailov2023direct, zhao2024towards, lee2024mechanistic}. In terms of classifying them by the parameters covered during training, they include full parameters fine-tuning \citep{eldan2023whos, jang-etal-2023-knowledge, yao2024large, rafailov2023direct}, sparse fine-tuning \citep{chang2023localization, stoehr2024localizing}, and parameter-efficient fine-tuning \citep{lu2024eraser, chen2023unlearn}. Additionally, there are also a few knowledge editing methods \citep{patil2024can}. We present the specific logic details of each method in \S\ref{appendix: unlearning methods}.

\paragraph{Knowledge Storation in Large Language Models}
Studying how knowledge is stored, transferred, and extracted in LLMs has always been an important direction in the research of LLM's interpretability \citep{meng2022locating, geva2021transformer, sukhbaatar2015end, geva2023dissecting}. It is known that in transformer-based language models, the MLP is a crucial component for storing the model's factual knowledge, and its sub-layers can be viewed as key-value memories \citep{geva2021transformer}. To be specific, the first layer\footnote{Currently, in most decoder-only models such as GPT-2 \citep{radford2019language} and GPT-J\citep{chen2021evaluating}, the MLP component has two layers, while in LLaMA \citep{touvron2023llama} it has three layers. However, we can still consider LLaMA's first two layers together as the key matrices, with their output serving as the coefficient scores.} of MLP sublayers can be viewed as a matrix $W_K$ formed by key vectors $\{\mathbf{k}_1, \mathbf{k}_2, \ldots, \mathbf{k}_n\}$, used to capture a set of patterns in the input sequence, and ultimately outputting the coefficient scores. The second layer can be viewed as a matrix $W_V$  formed by value vectors $\{\mathbf{v}_1, \mathbf{v}_2, \ldots, \mathbf{v}_n\}$, with each value vector containing the corresponding factual knowledge (represented through token distributions). Finally, the MLP's output can be defined as the sum of value vectors weighted by their memory coefficients: %, as shown in Eq.~(\ref{eq:mlp}):
\begin{equation}
\label{eq:mlp}
    \mathbf{M}^\ell = f \big( W^\ell_K \mathbf{x}^\ell \big) W^\ell_V = \mathbf{m}^\ell W^\ell_V, 
\end{equation}
where $\mathbf{M}^\ell$ represents the output of the MLP in the transformer's $\ell$-th layer for an input hidden state $\mathbf{x}^\ell$ at that layer with the parameters, $W^\ell_K$ and  $W^\ell_V \in \mathbb{R}^{n\times d}$.  $f$ is a non-linearity function\footnote{Here, the bias term is omitted for brevity.}. $\mathbf{m}^\ell\in \mathbb{R}^{n}$ represents the coefficient scores. The dimension size of hidden states is $d$ and it is $n$ for the intermediate MLP.

In addition to the MLP, primarily responsible for knowledge storage, the attention component is currently considered the main component responsible for knowledge transfer and extraction in language models~\citep{geva2023dissecting}.
Here, we will not go into detail about its specific structure but only study the impact it has on knowledge extraction. The final computation formula for the hidden states in the language model is defined as: %shown in Eq.~(\ref{eq:hs}):
\begin{equation}
\label{eq:hs}
    X^{\ell+1} = X^\ell + \mathbf{M}^\ell + \mathbf{A}^\ell, 
\end{equation}
where $X^\ell$, $\mathbf{M}^\ell$ and $\mathbf{A}^\ell$ represent the hidden states, MLP's output, and the attention component's output in the transformer's $\ell$-th layer, respectively.

\section{Patching Investigation}
\label{sec: patching}

\paragraph{Hypothesis and Experimental Design}

Based on Eq.~(\ref{eq:mlp}) and Eq.~(\ref{eq:hs}), we hypothesize that there are three main reasons why the current fine-tuning-based unlearning methods appear successful in behavioral tests and seem to suggest that true unlearning has been achieved:

\begin{compactenum}
  \item The coefficients $\mathbf{m}^\ell$ are changed after fine-tuning, leading to a change in the activations of the MLPs;
  \item The value vectors $W^\ell_V$ in MLPs are changed, causing a change in the knowledge they contain; 
  \item The change that happens in attention components caused the model's focus and the corresponding information extracted by these attention components $\mathbf{A}^\ell$ to change, thus reducing the target knowledge-related information in the output. 
\end{compactenum}

Here, for the sake of simplicity and better understanding, we continue to use the definitions of $\mathbf{m}^\ell$, $W^\ell_V$, and $\mathbf{A}^\ell$ as given in Eq.~(\ref{eq:mlp}) and Eq.~(\ref{eq:hs}) in the following.  We ignore the minor effects caused by other components or parameters, such as the language model's unembedding matrix and the normalization layers. Based on the possible reasons described above, on the unlearned model, we conduct three different sets of activation patching or components' parameter restoration experiments, trying to recover the output of the target knowledge in the unlearned model. The specific operation process is as follows:

\begin{compactenum} 
  \item In the first set of experiments, we restore the coefficient scores $\mathbf{m}^\ell$ corresponding to each MLP component, layer by layer, in the language model, without making any intentional changes to the value vector parameters $W^\ell_V$ of the MLPs or the attention components' states $\mathbf{A}^\ell$ in any layer. 
  \item In the second set of experiments, we restore the parameters of value vectors $W^\ell_V$ in MLPs layer by layer, recovering the knowledge they originally contained. In this process, we avoid making intentional changes to the unlearned model's original coefficients $\mathbf{m}^\ell$ and the attention components' states $\mathbf{A}^\ell$. 
  \item In the third set of experiments, we restore the original attention components' states $\mathbf{A}^\ell$, but without intentionally altering the MLPs' coefficient scores $\mathbf{m}^\ell$ or the value vectors' parameters $W^\ell_V$, only studying the impact brought by the attention components which are responsible for extracting and transferring knowledge.
\end{compactenum}

To evaluate the extent of knowledge restoration, we propose the metric of \textbf{Knowledge Recovery Score (KRS)}:
\begin{equation}
    \label{eq:krs}
 KRS  = 1 - \mathrm{loss}^{*o}/\mathrm{loss}^*,
\end{equation}
% Should be MSE loss!
where the losses are the average of $MSE(\cdot)$ on $L_{i,n}^*$ and $L_{i,n}$ and on $L_{i,n}^{*o}$ and $L_{i,n}$, respectively.  $MSE(\cdot)$ represents the mean squared error (MSE) loss function. $L$, $L^*$, and $L^{*o}$ are the logit distribution of the subsequent token produced by the vanilla model, unlearned model, and unlearned-then-recover model, respectively.  The average loss is computed on the next $I$ generated tokens on $N$ knowledge-related questions.
%defined as:  
\if 0
\begin{equation*}
  \mathrm{loss}^*  = \frac{1}{N*I} \sum_{n=1}^{N} \sum_{i=1}^{I} NLL( S(L_{i,n}^*), S(L_{i,n})),  
\end{equation*}
\begin{equation*}
\mathrm{loss}^{*o}  = \frac{1}{N*I} \sum_{n=1}^{N} \sum_{i=1}^{I} NLL( S(L_{i,n}^{*o}), S(L_{i,n}) ),
\end{equation*}  
\fi

Finally, if KRS approaches 1, it indicates $L_{i,n}^{*o}$ and $L_{i,n}$ that are nearly consistent, representing a higher degree of knowledge recovery. Conversely, a lower KRS suggests a lower degree of that.

%\subsection{Activation Patching and Parameters Restoration Experiments}
\paragraph{Activation Patching and Parameters Restoration Experiments}

We conduct the experiments on two recent LLMs, LLaMA2-7B-chat \citep{touvron2023llama} and OLMo-7B \citep{groeneveld2024olmo}.  
% To xx, 
We apply two example finetuning-based unlearning methods, DPO~\cite{rafailov2023direct} and Gradient Difference~\citep{yao2024large}, to perform unlearning on the large language models and calculate the average KRS scores. Inspired by~\citep{eldan2023whos}, which tries to unlearn the concept knowledge of ``Harry Potter'' in language models, we extend this experiment by selecting 10 well-known concepts per model from the ConceptVectors Benchmark~\citep{hong2024intrinsicevaluationunlearningusing}, which is a collection of concepts that language models are well-acquainted with and have substantial knowledge about. Examples of them are provided in Table~\ref{tab:example data} of \S\ref{appendix: corpus}.
For the unlearning training, we use the texts containing the corresponding concepts from Redpjama\footnote{https://www.together.ai/blog/redpajama} and Dolma~\citep{soldaini2024dolma}. Redpjama is a replication of the pretraining corpus for the LLaMA model, while Dolma is the open-source pre-training dataset for the OLMo model. Detailed information is provided in \S\ref{appendix: corpus}. So here we can ensure that the knowledge to be unlearned was at least seen by the model during the pre-training process, 
and that the training data used more broadly covers the textual sources from which the model acquired the corresponding knowledge about certain concepts.

After obtaining the unlearned model, we follow the steps mentioned in the hypothesis to perform activation patching and parameter restoration experiments on the unlearned models. To calculate the Knowledge Recover Scores, we set $I$ to 30 and $N$ to 10, indicating the generation of the next 30 tokens and the selection of 10 questions related to each concept. To make the recovery effects more pronounced and the whole process easier to observe, we adopt techniques from \citep{meng2022locating, meng2023massediting} which implemented causal mediation, setting the size of the recovery window to five. This allows us to observe the average effects of recovering five consecutive layers at a time. Details can be found in \S\ref{appendix: corpus}.

\begin{figure}[t]
    \centering
    \includegraphics[scale=0.280]{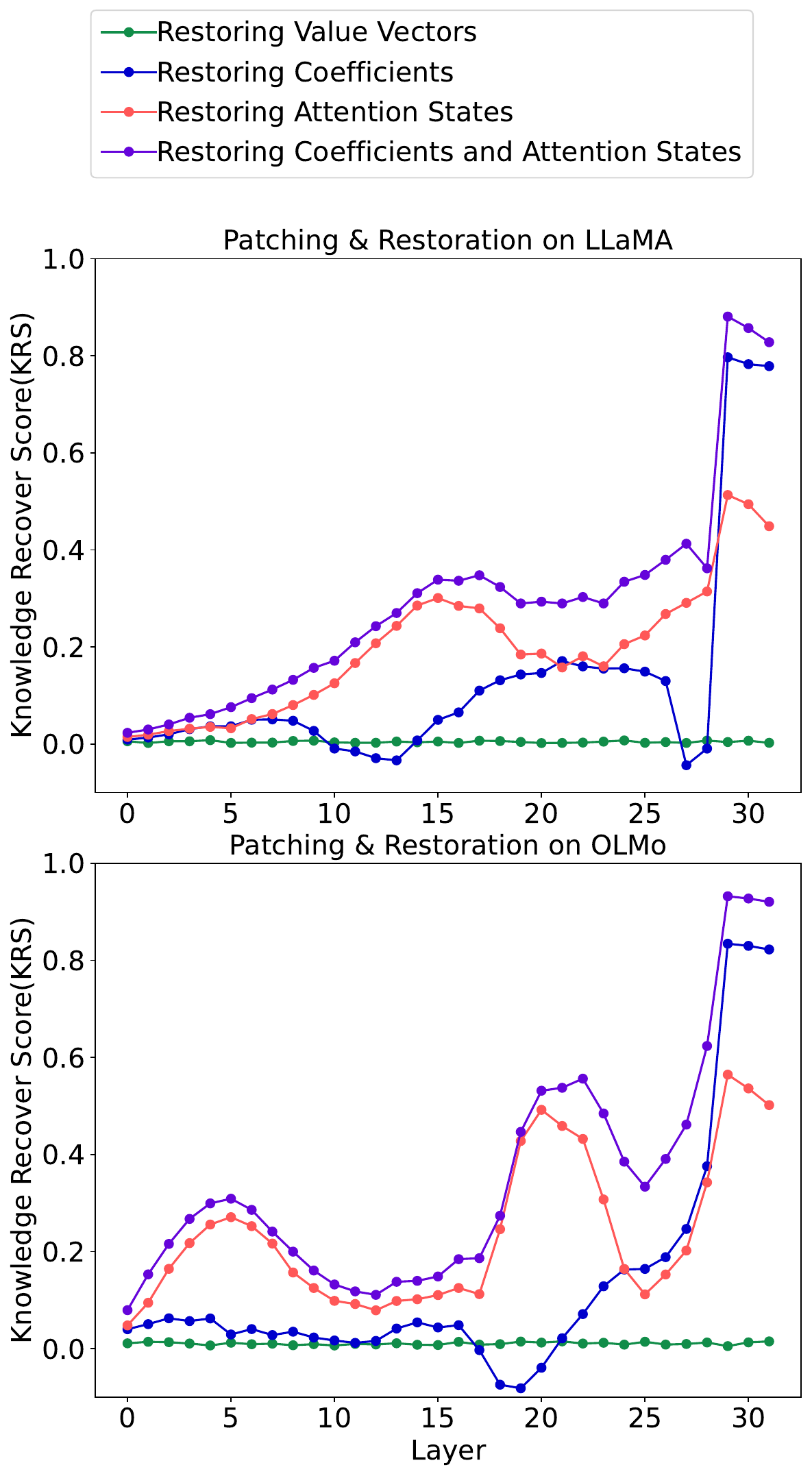}
    \caption{Results of KRS on LLaMA and OLMo under three activations patching or parameters restoration settings. We also included another setting that restores both attention and coefficients to compare the final outcomes.}
    \label{fig: patching_fig}
\end{figure}

The specific results are shown in Fig.~\ref{fig: patching_fig}. From our analysis, surprisingly, we observe that when we solely recover the parameters contained in the value vectors of each layer in the unlearned model without interfering with the coefficients or attention components' states, the recovery of the target knowledge is negligible (The KRS scores are all below 0.001). This holds regardless of which layer is recovered, and regardless of the specific model being considered.

However, when recovering the attention components' states in the intermediate layers (from the 15th layer onward) or deeper layers (from the 27th layer onward), we can observe that the average KRS for both models has increased to exceed 0.3 and 0.4, respectively, indicating that a significant portion of the corresponding knowledge has been recovered. What's more, restoring the coefficients of the MLPs in the intermediate layers (from the 20th layer onward) and deeper layers (from the 29th layer) also yields impressive knowledge recovery effects.

The layers at which the scores start to increase under the two settings generally align closely with the observation by~\citet{geva2023dissecting} that the MLP modules recall knowledge in intermediate layers, and the attention components mostly start to extract and transfer information in the deeper layers. or after the model has completed the relevant knowledge recall. We also tried simultaneously recovering the coefficients and attention states and found that the model can achieve much greater knowledge recovery, with the peak KRS score exceeding 0.9 on both models.

Additionally, it is noteworthy that, simply restoring the coefficient scores of the MLP outputs from the last two or three layers can significantly elevate the KRS of the unlearned LLaMA and OLMo models to 0.8 or above. This suggests that the coefficient scores of the MLPs in the last layers might play a crucial role in the final behavior results of the LLM. To better isolate the effects of restoring $\mathbf{m}^\ell$, $W^\ell_V$, and $\mathbf{A}^\ell$ individually and support the above argument, we present a more rigorous patching and restoration experiment in \S\ref{appendix: rigorous}, with the corresponding results shown in Figure~\ref{fig: patching_rigorous}. Ultimately, we found that the restoration of the attention states also contributed to the coefficients of the MLP in the final layers, further confirming that these coefficients carry the primary role of achieving the effects of finetuning-based unlearning. 
It also indicates that fine-tuning largely adjusts the model's behavior by modifying the coefficients of the deep MLP layers, likely because this enables faster adaptation compared to other knowledge adjustment mechanisms, such as altering knowledge encoded in the MLP itself. 
This phenomenon and the potential defensive strategy have not been discussed in the previous literature, warranting further investigation in future studies.

Overall, these results all further confirm that the finetuning-based unlearning methods essentially do not modify the model knowledge contained in the value vectors, but adjust the way knowledge is called during the fine-tuning process, either by adjusting the coefficients to modulate the MLP activation or by adjusting the attention to extract and transfer knowledge.

% we follow the setting in \citet{meng2022locating}

\section{Global Negative Effect of Fine-Tuning Unlearning}
\label{sec: unlearning experiments}

In the previous section, we demonstrated that these finetuning-based methods alter the model's final behavior by adjusting the MLP output coefficients in the final layers. Therefore, we hypothesize that this behavioral change will has a global effect, potentially impacting the output of unrelated knowledge as well. In this section, we verify this hypothesis through the following experiments.

\begin{figure}[t]
    \centering
    \includegraphics[scale=0.20]{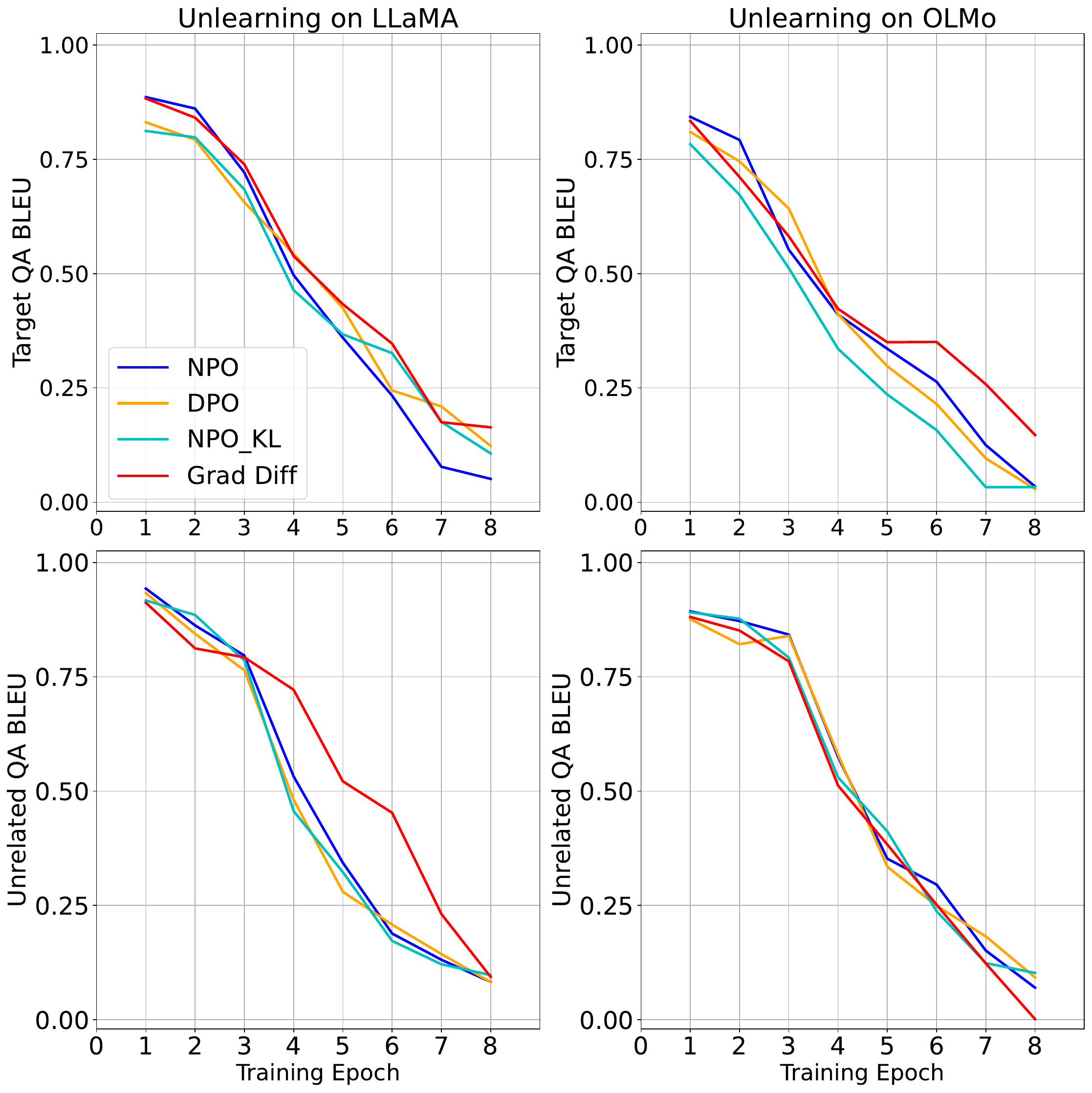}
    \caption{Unlearning testing results on LLaMA and OLMo for each training epoch.}
    \label{fig: unlearning_qa_bleu}
\end{figure}

We apply four fine-tuning-based unlearning methods to the concepts used in \S\ref{sec: patching} on their pretraining text sources (from RedPajama and Dolma) with the goal of erasing the learned knowledge during pretraining through a reverse process. These methods are as follows:
DPO~\citep{rafailov2023direct}, NPO~\citep{zhao2024towards}, NPO+KL~\citep{zhao2024towards} and Gradient Difference~\citep{yao2024large}.
% We apply the behavioral testing criteria to evaluate the performance of the finetuning-based unlearning methods, to specifically study what kind of impact these methods would have on the model's general or unrelated capabilities. 
% In the following, we test five fine-tuning-based methods: DPO~\citep{rafailov2023direct}, NPO~\citep{zhao2024towards}, NPO+KL~\citep{zhao2024towards}, Gradient Difference~\citep{yao2024large}, and Gradient Ascent~\citep{jang-etal-2023-knowledge}.  
% For each method, we tune the hyperparameters and obtain three different data points with varying performance. 
% Additionally, for better comparison, we adopt a knowledge editing method, MEMIT~\citep{meng2023massediting}, along with two different loss functions suitable for unlearning~\citep{patil2024can}.
The details of these baselines and data statistics are shown in \S\ref{appendix: unlearning methods} and \S\ref{appendix: corpus}.
We evaluate the unlearning effectiveness of these methods on the concepts' related QA pairs and the unlearning impact on unrelated QA pairs, reporting the average scores of \textbf{BLEU}~\citep{papineni2002bleu} by comparing the model's response before and after unlearning. In Figure~\ref{fig: unlearning_qa_bleu}, we report their performance at the end of each training epoch respectively.

We can observe that for finetuning-based methods, as the number of training epochs increases, aiming to achieve a lower target QA BLEU score, the corresponding unrelated QA BLEU score also decreases accordingly, exhibiting a positive correlation.  
This suggests that the impact of finetuning-based methods on the model's output behavior is global. While unlearning the target knowledge, they inadvertently alter the output behavior or manner for unrelated knowledge to a certain degree. 
% In contrast, MEMIT, which selectively modifies the MLP parameters, can mitigate this correlation to some extent. Even when unlearning the target knowledge to a very low level, it can still largely preserve the model's output behavior and performance on unrelated knowledge, maintaining it close to its original state.

\section{Discussion and Conclusion}

We have deeply investigated the reasons why fine-tuning-based unlearning methods seemingly succeeded in behavior-based testing for large language model unlearning: Through activation patching and parameter restoration experiments, we find that these methods alter the way knowledge is extracted by changing MLP activations or model's attention, ultimately affecting the output. This is evidenced by the fact that the model's output regarding the target knowledge is largely restored after patching the activations and the attention components' states. Furthermore, we conduct experiments on the pretraining datasets of two models, to test the models' capabilities after unlearning, verifying that in addition to unlearning the corresponding knowledge, fine-tuning-based methods that by altering the way the model accesses knowledge, will significantly impair the model's other unrelated capabilities, causing a certain degree of capability degradation.

\section{Limitations}

In the experiments detailed in \S\ref{sec: patching}, we have disregarded the potential unlearning impact caused by parameter changes in other model components during the fine-tuning process. This decision is based on the observation that the impact of such changes appears to be minimal.  For instance, during our parameter comparison analysis, we found that the changes in the unembedding matrix and normalization layer parameters resulted in cosine similarity values above 0.999. This suggests that the modifications to these components are quite small in magnitude.

However, it remains unclear whether even such minimal parameter changes can still have any meaningful effect on the model's overall behavior and knowledge. Further verification and analysis would be needed to conclusively determine the extent to which these ancillary parameter updates might influence the unlearning outcome.

\section*{Acknowledgements}
%.....
The work was fully supported by the IDEA Information and Super Computing Centre (ISCC), National Natural Science Foundation of China (Grant No. 62406114), the Guangzhou Basic and Applied Basic Research Foundation (Grant No. 2023A04J1687), and the Fundamental Research Funds for the Central Universities (Grant No. 2024ZYGXZR074).
Di Wang and Lijie Hu are supported in part by the  funding BAS/1/1689-01-01, URF/1/4663-01-01,  REI/1/5232-01-01,  REI/1/5332-01-01, and URF/1/5508-01-01 from KAUST, and funding from KAUST - Center of Excellence for Generative AI, under award number 5940.

\bibliography{custom}

\appendix

\section{Details in Existing Unlearning Methods}
\label{appendix: unlearning methods}

In this section, we provide a more detailed introduction to the LLM unlearning methods we used in \S\ref{sec: patching} and \S\ref{sec: unlearning experiments}.

\begin{itemize}
    % \item \textbf{Gradient ascent}~\citep{jang-etal-2023-knowledge}, a simple and widely adopted unlearning method, maximizes the next-token prediction loss over a set of text sequences that we wish the LLM to forget, thereby ``revert'' the optimization on the forget set via gradient descent during pretraining.
    \item \textbf{Gradient Difference}~\citep{yao2024large}, based on Gradient Ascent, it adds a regularization term to minimize the KL divergence between the unlearned and the original LLM on a reference text dataset, thus preventing the model from catastrophic deterioration of its general capability.
    \item \textbf{Direct Preference Optimization (DPO)}~\citep{rafailov2023direct}, which maximizes the log-likelihood ratio between generating the preferred and the unfavored responses, while retaining a small shift from the original LLM predictive distribution,
    \item \textbf{Negative Preference Optimization (NPO)}~\citep{zhao2024towards}, which discards the favored responses and only minimizes the prediction probability of the unfavored answers.
    \item \textbf{NPO+KL} which adds to NPO a KL divergence loss between the model's outputs before and after unlearning.
    % \item \textbf{MEMIT+Empty}. \textbf{MEMIT} is a prominent model editing algorithm. We follow \citet{patil2024can}, who have proposed multiple methods to adapt MEMIT from knowledge editing to knowledge removal, and set the new target in the editing task to a ``dummy'' meaningless object.
    % \item \textbf{MEMIT+Entropy}. We follow \citet{patil2024can} and replace the original objective of MEMIT with a new objective that suppresses tokens related to the object from appearing with high probability in the vocabulary projection of hidden representations during inference. This is achieved by maximizing the entropy of the next-token probability distribution over the vocabulary for every layer.

\end{itemize}

\section{Unlearning Experiment's Corpus}
\label{appendix: corpus}

Here, we present detailed information about the data used for activation patching experiments and the unlearning experiments conducted in \S\ref{sec: patching} and \S\ref{sec: unlearning experiments}. We select 10 well-known concepts from ConceptVectors Benchmark~\citep{hong2024intrinsicevaluationunlearningusing} and extract 6,000 corresponding training data segments containing knowledge about the respective concepts per model from the pre-training datasets of Redpjama and Dolma. These extracted data segments are used for unlearn training of the two models respectively. For each concept, we also include ten related questions from the ConceptVectors Benchmark, along with 50 unrelated questions sampled from other unrelated concepts. These are used in \S\ref{sec: unlearning experiments} to evaluate the unlearning effectiveness from the behavior perspective on the specific concepts, as well as to assess whether the model's unrelated capabilities were affected. We have manually checked and verified that the vanilla LLaMA and OLMo models can accurately answer these selected questions, indicating that the models possess the knowledge.
All the statistics and examples are shown in Table~\ref{tab:training data} and Table~\ref{tab:example data}, respectively.

\begin{table*}[t]
\centering
\resizebox{\linewidth}{!}{
\begin{tabular}{lccccc}
\toprule
Data Sources & \# selected concepts & \# of paragraphs per concept & \# of words per paragraph & \# of QA pairs & \# of unrelated QA pairs \\
\midrule
Redpjama & 10 & 6000 & 1514.65 & 20 & 50 \\
Dolma & 10  & 6000 & 2261.25 & 20 & 50\\
\bottomrule
\end{tabular}
}
\caption{Statistics of the training data for the unlearning experiments on LLaMA and OLMo}
\label{tab:training data}
\end{table*}

\begin{table*}[t]
\setlength{\belowcaptionskip}{-10pt}
\setlength\tabcolsep{4.0pt}
\centering

% \vspace{10px}
\resizebox{\linewidth}{!}{
\begin{tabular}{p{4cm}p{6.5cm}p{5cm}p{5cm}}
% \toprule
\textbf{Concept} &\textbf{Training Data Snippets} &\textbf{Example QA} & \textbf{Example Unrelated QA}  \\
\midrule
Harry Potter(LLaMA) & Harry Potter is a series of seven fantasy novels written by British author J. K. Rowling. The novels chronicle the lives of a young wizard, Harry Potter, and his friends Hermione Granger and Ron Weasley, all of whom are students at Hogwarts School of Witchcraft and Wizardry.. & Who is the author of the Harry Potter book series?
        
What is the name of the first book in the Harry Potter series?.. & In which century did William Shakespeare live and write?
          
What town is traditionally considered Shakespeare's birthplace?.. \\
Star Wars(LLaMA) & Star Wars is an American epic space opera media franchise created by George Lucas, which began with the eponymous 1977 film and quickly became a worldwide pop culture phenomenon.. & Who is Darth Vader's son?

What is the weapon used by Jedi Knights?.. & What are the twelve zodiac signs?

Which astrological sign is represented by the lion?..     \\
Amazon Alexa(LLaMA) & Amazon Alexa or Alexa is a virtual assistant technology largely based on a Polish speech synthesizer named Ivona, bought by Amazon in 2013. It was first used in the Amazon Echo smart speaker and the Echo Dot, Echo Studio and Amazon Tap speakers developed by Amazon Lab126.. & What year was the Amazon Alexa Voice Assistant first introduced to the public?

What are some of the primary functions of Amazon Alexa Voice Assistant?..  & Who betrayed Jesus to the authorities in the Bible?

What is the longest book in the Bible in terms of chapters?..\\

\midrule
Ebay(OLMo) & eBay Inc. ( EE-bay, often stylized as ebay) is an American multinational e-commerce company based in San Jose, California, that brokers customer to customer and retail sales through online marketplaces in 190 markets worldwide.. & What is the name of Japan's most popular boy band?

Who is Japan's most famous anime creator? ..& 
What does IRC stand for?

When was IRC first developed?..\\
Olympic Games(OLMo) &The modern Olympic Games or Olympics (French: Jeux olympiques) are the leading international sporting events featuring summer and winter sports competitions in which thousands of athletes from around the world participate in a variety of competitions.. & 
When were the first modern Olympic Games held?,
            How often are the Summer Olympics held?.. & What is virtual reality?
            
 How does virtual reality technology work?..\\
Diabetes(OLMo) & Diabetes mellitus, often known simply as diabetes, is a group of common endocrine diseases characterized by sustained high blood sugar levels. Diabetes is due to either the pancreas not producing enough insulin, or the cells of the body becoming unresponsive to the hormone's effects..
& What is diabetes?
          
          What are the main types of diabetes?.. &  What is the capital city of Pakistan?
          
          What is the currency of Pakistan?.. \\

\bottomrule

\end{tabular}}
\caption{Example extracted data from the Redpjama and Dolma pre-training datasets.} 
\label{tab:example data}
\end{table*}

\section{More Rigorous Patching Investigation}
\label{appendix: rigorous}

In \S\ref{sec: patching}, during our activation patching and parameters restoration experiments, we restore $\mathbf{m}^\ell$, $W^\ell_V$, or $\mathbf{A}^\ell$ layer by layer respectively, while avoiding intentional changes to the other two states in the unlearned model. However, for instance, restoring $\mathbf{A}^\ell$ in $\ell$-th layer may aid in the recovery of $\mathbf{m}^\ell$ in subsequent layers, ultimately leading to an improvement in KRS. Therefore, in this part of the experiment, when restoring each element layer by layer, we purposefully keep the other two elements unchanged (e.g., when restoring $\mathbf{A}^\ell$, we maintain the original states of $\mathbf{m}^\ell$ and $W^\ell_V$ for both the current and subsequent layers). This approach thoroughly isolates the effects of these three different elements.

\begin{figure}[t]
    \centering
    \includegraphics[scale=0.280]{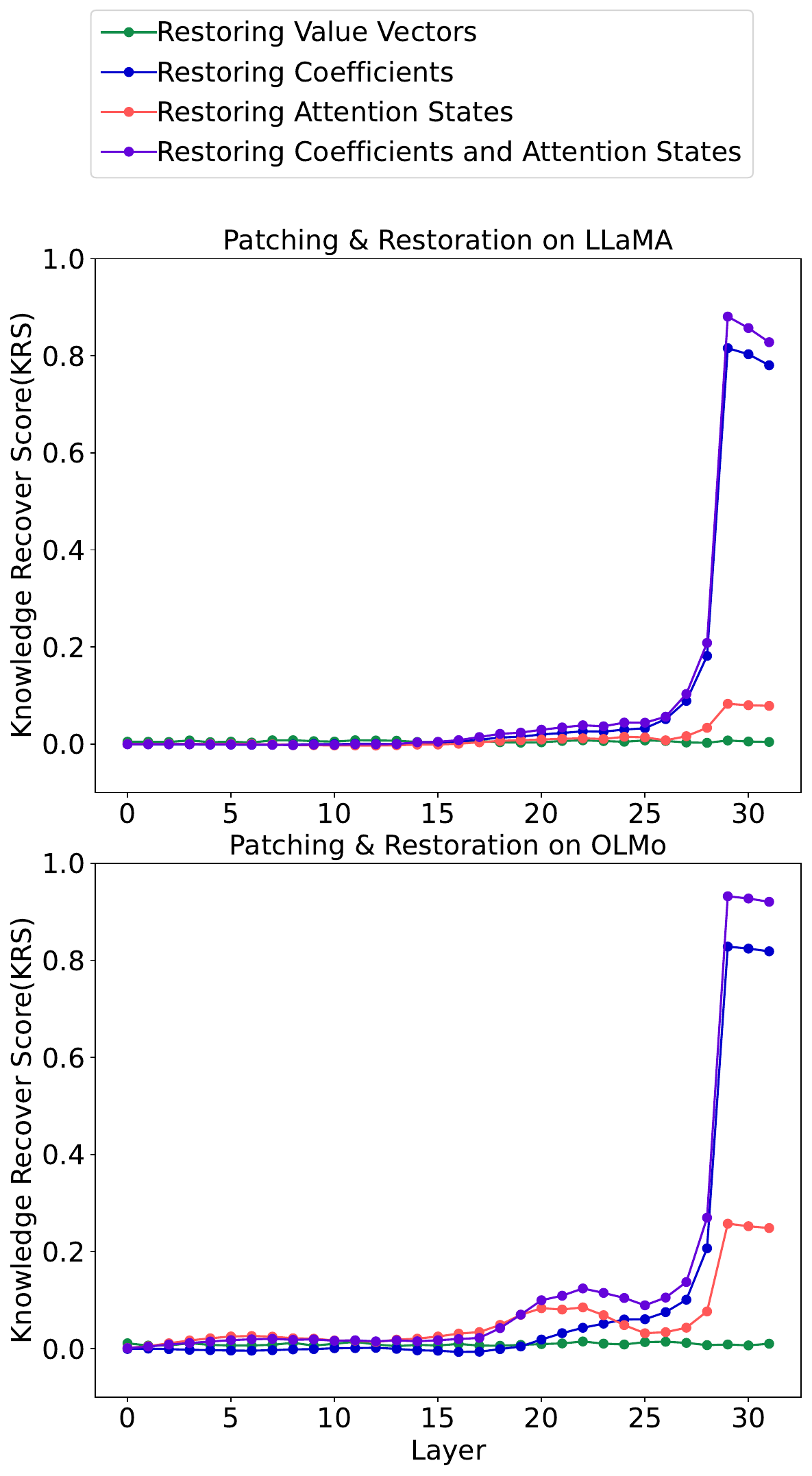}
    \caption{Results of KRS on LLaMA and OLMo under three activations patching or parameters restoration settings, isolating the effects of the two others when investigating each factor individually. }
    \label{fig: patching_rigorous}
\end{figure}

Figure~\ref{fig: patching_rigorous} presents the results in this setting. We can observe the following: (1) When $W^\ell_V$ is restored layer by layer, its effect on improving KRS remains very small, which is consistent with prior experiments. (2) When restoring $\mathbf{A}^\ell$ layer by layer and isolating its effects from the other two factors, its contribution to KRS remains insignificant, staying at a low level and only increasing to around 0.08 on LLaMA and 0.2 on OLMO in the final layers. (3) When $\mathbf{m}^\ell$ is restored layer by layer, isolating its influence from the other elements, we observe a notable rise in KRS in the last three layers, reaching values as high as 0.8 or above. This supports the idea that neurons responsible for $\mathbf{m}^\ell$ in the MLP components of the final layers primarily carry the unlearning effects of these finetuning-based methods.

\end{document}